\documentclass[a4paper]{article}

\usepackage{amsmath}
\usepackage{amsthm}
\usepackage{amsfonts}
\usepackage{amssymb}

\usepackage{lipsum}
\usepackage{booktabs}

\usepackage{setspace}

\usepackage{enumerate}
\usepackage{enumitem}
\usepackage{colonequals}
\usepackage[all]{xy}
\usepackage{todonotes}
\usepackage{algorithm}
\usepackage{algpseudocode}

\newcommand{\R}{\mathbb R}

\usepackage{graphicx}

\newcommand{\best}[1]{\textbf{#1}}

\newtheorem*{theorem*}{Theorem}{\bf}{\it}

\newtheorem*{lemma*}{Lemma}{\bf}{\it}

\newtheorem*{proposition*}{Proposition}{\bf}{\it}

\theoremstyle{definition}

\newtheorem*{definition*}{Definition}
\theoremstyle{remark}

\numberwithin{equation}{section}
\usepackage{palatino, url, multicol}

\usepackage{tikz}

\title{A visual observation on the geometry of UMAP projections of the difference vectors of antonym and synonym word pair embeddings}
\author{$\R$ami Luisto\thanks{%
  Faculty of Information Technology, University of Jyv\"askyl\"a, Jyv\"askyl\"a;
  HUS Helsinki University Hospital, Helsinki;
  Digital Workforce Services, Helsinki.
  \texttt{rami.m.luisto@jyu.fi}}}

\begin{document}

\maketitle

\begin{abstract}
  Antonyms, or opposites, are sometimes defined as \emph{word pairs that have all of the same contextually relevant properties but one}. Seeing how transformer models seem to encode concepts as directions, this begs the question if one can detect ``antonymity'' in the geometry of the embedding vectors of word pairs, especially based on their difference vectors. Such geometrical studies are then naturally contrasted by comparing antonymic pairs to their opposites; synonyms.

  This paper started as an exploratory project on the complexity of the systems needed to detect the geometry of the embedding vectors of antonymic word pairs. What we now report is a curious ``swirl'' that appears across embedding models in a somewhat specific projection configuration.
\end{abstract}

\section{Introduction}

Studying antonyms and synonyms in NLP has a long and rich research history, predating transformer architecture by decades, see e.g.\ \cite{Nguyen2016Contrast} and the references within. In this paper our focus is on the geometry of the pairs of word embeddings of antonymic or synonymic word pairs.
For us a very fitting point of view is in \cite[p.\ 170]{murphy2003semantic}, where Murphy defines antonyms as word pairs that ``\emph{[...] have all of the same contextually relevant properties but one.}'' This is naturally contrasted with how in the field of AI, modern NLP models model language as collections of high-dimensional vectors (see e.g.\ \cite{bolukbasi2016man,vaswani2017attention}). In particular, it seems that \emph{concepts} are mapped to various \emph{directions} in these high-dimensional spaces; \cite{bolukbasi2016man,elhage2022toy}. 

From this point of view we are tempted to try and paraphrase Murphy's definition of antonyms as ``Word pairs that agree on all contextual \emph{dimensions} but one.'' The question then is whether this single ``semantic dimension'' would correspond in any (detectable) way to a Euclidean dimension in the embedding spaces of any modern language models. When taken literally, the literature provides a very thorough ``no'' as an answer; see e.g.\ \cite{ali2019antonym,glavavs2018discriminating,samenko2021intuitive,xie2021mixture}.
For example \cite{samenko2021intuitive} shows that antonyms and synonyms can be differentiated with a high accuracy from their word2vec, GloVe or FastText embeddings, but the contrastive map they build is very much non-linear. On the other hand \cite{xie2021mixture} suggest that the differentiating boundary between antonyms and synonyms is not a single hyperplane but a collection of local nonlinear boundaries: ``\emph{Two hypotheses underlie our method: (a) antonymous words tend to be
similar on most semantic dimensions but be different on only a few salient dimensions; (b) the salient dimensions may vary significantly for different antonymies throughout the whole distributional semantic space.}'' But while a simple Euclidean direction for depicting antonymity seems not to exist in modern language models, this still leaves open the question on what kind of geometry in the embedding space might correspond to the concept of antonymity.

It is natural here to compare antonyms with synonyms in particular. Synonyms are simply words that mean (roughly) the same thing, and in particular we could rephrase Murphy's definition of antonyms from above as ``A pair of words which are synonyms except for one relevant property.'' This is in contrast to most other word pairs which will typically differ by \emph{several} relevant properties. In particular, we would expect that if we study pairs of embedding vectors coming from synonyms, antonyms, random word pairs, the levels of similarity should be decreasing.

The purpose of this paper is to report a very particular observation from a large set of experiments. We set out to take the Stuttgart dataset \cite{nguyen2017distinguishing} containing antonymic and synonymic word pairs and study the geometries of the embedding vector pairs under various transformer models.
With the embedding vectors we studied both their differences and their concatenations, and ran the resulting point clouds through various projection methods like PCA, t-SNE and UMAP. We also tested a large collection of various classification algorithms to see how their performance differed on the task. What we report here is a visual observation in the UMAP projections of the difference vectors, especially when the dataset was extended with certain ``control pairs'' of words we describe later in Section \ref{sec:data}. One should be suspicious here if we are simply reporting a result of $p$-hacking\footnote{I.e.\ if we just generated a large amount of projection configurations and report the one combination that happened to randomly show some geometrical artefacts.}, but as we note later on the effect seems to persist across UMAP parametrizations and across very different types of embedding models.

Besides our report on ``the swirl'' we observe, we also note that the projections can be used to produce a transductive classifier that performs on par with SOTA methods in the Antonym Synonym Differentiation (ASD) task, though with some caveats. We discuss this briefly in Section \ref{sec:classification}.

The source code for this paper and the scripts that can be used to replicate the Swirl are available at \url{https://github.com/ramiluisto/CuriousSwirl.git}.


\subsection{A few notes on some related literature}
\label{subsec:litreview}

We aim to keep this a short visual note, but we give a quick walk-through of some relevant literature for the benefit of a curious reader.

Many of the earlier NLP systems, especially those related to word embeddings, relied on co-occurrence statistics for word comprehension, which made synonyms and antonyms tricky since their co-occurrence patterns can be quite similar, see e.g.\ \cite{brunila2022company}.
In a more ``manual'' approach, \cite{Ono2015, Nguyen2016Contrast,  Mrksic2016Counterfit, Mrksic2017AttractRepel, Dou2018AntSent} used methods where they purposefully pushed known antonyms apart.
General neural network approaches proved also fruitful in the ASD task, see \cite{Nguyen2017AntSynNET,Etcheverry2019Siamese,Ali2019Distiller}.

The \emph{Intuitive Contrasting Map} of \cite{samenko2021intuitive} applies a light non-linear
projection to classical word embeddings (like word2vec and GloVe) and shows that synonym and antonym cosine distributions
separate cleanly after transformation, indicating that antonymy is
already latent in the geometry. In the domain of transformers, attention-head probing has revealed that specific BERT heads systematically connect semantically related word pairs, including both synonyms and antonyms, independently of context \cite{Serina2023Heads}.



Finally we note that we are studying a topic in the interaction of linguistics and NLP. For the more linguistic side of things we direct the reader to \cite{koptjevskaja2024cross} and the references within. See also \cite{hippisley2016cambridge} section on antonyms or \cite[Chapter 5]{murphy2003semantic}.

\section{Data and Methods}
\label{sec:data}

Our primary data source is ``The Stuttgart dataset'', \cite{nguyen2017distinguishing}, containing a large collection of synonym and antonym pairs, divided into adjective pairs, noun pairs and verb pairs. For each word we construct their embedding vectors with one of our embedding models. 

We originally focused only on context-free embeddings of transformers, but expanded our approach to different types of embedding models to better understand how typical it is for the observed geometry to appear. For this report we use the ``classical'' word2vec and GloVe, the BERT context-free embeddings and two proprietary models from OpenAI used through an API: text-embedding-3-small and text-embedding-3-large.

We remark that the text-embedding-3-small/large models are general text-embedding models and produce embeddings for any input text, and thus from there we get embeddings for the full set of words found in the Stuttgart dataset. Word2Vec and GloVe are static embedding models, though they seem to have a good coverage of the words we study here. For the BERT model we use the context-free embedding vectors, i.e.\ we take those words that are single tokens in the BERT tokenizer and take the corresponding context-free embedding vectors. Details of this coverage and the embedding dimensions of the models are listed in Table \ref{tab:model-coverage}.

\begin{table}[ht]
  \centering
  \caption{Model coverage: vocabulary size and embedding dimensionality.}
  \label{tab:model-coverage}
  \begin{tabular}{lrc}
  \toprule
  Model & Words & Dim \\
  \midrule
  Word2Vec         & 9{,}056 & 300 \\
  GloVe            & 9{,}250 & 300 \\
  BERT Base        & 4{,}783 & 768 \\
  text-embedding-3-small & 9{,}405 & 1{,}536 \\
  text-embedding-3-large & 9{,}405 & 3{,}072 \\
  \bottomrule
  \end{tabular}
\end{table}

One of the original aims of this project was to also train various classifiers to differentiate between antonyms and synonyms. We were concerned that the base words in the antonym and synonym pairs might have underlying differences, and so we also created two synthetic datasets; ``shuffled antonyms'' and ``shuffled synonyms'' in which we would take the set of words appearing in e.g.\ antonym pairs in either position, and sample from these new word pairs randomly. Thus any classifier classifying antonyms and synonyms could be viewed in the context of how well it performed in classifying between shuffled antonym pairs and shuffled synonym pairs. These control sets turned out to be very interesting in our visualization experiments.

\begin{figure}[htbp]
  \centering
  \includegraphics[width=\textwidth]{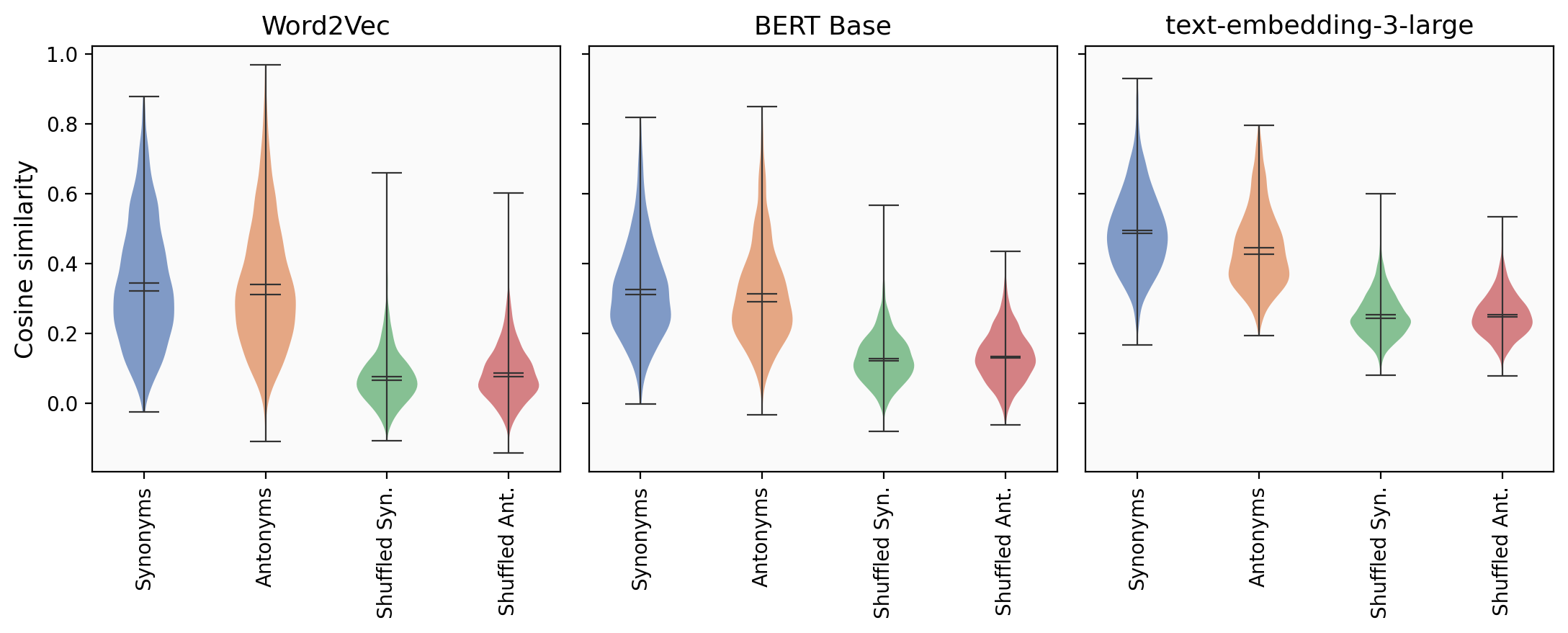}
  \caption{Cosine similarity distributions between word pairs within each dataset for four embedding models.}
  \label{fig:cosine-sim-dist}
\end{figure}

Figure \ref{fig:cosine-sim-dist} shows the distribution of cosine similarities between the words of each pair in our four datasets. for three of the models we use here. We note that at this level, the synonym and antonyms datasets resemble each other quite much, as do the two shuffled datasets, thought there are some small variations.

As a challenge with the data we note that as in \cite{nguyen2017distinguishing}, we aim to study a sort of \emph{binary} division of words. Is a pair of words antonyms or not? Are they synonyms or not? For many situations, a scale of antonymity or synonymity might be better. For example we consider the pairs (``big'', ``small''), (``huge'', ``small'') to be both antonym pairs, but in some sense one is more antonymic than the other. Here we limit ourselves just to binary division, but want to note that this is at times an ill-defined difference.

We study each of our five models separately. With our four datasets; an\-to\-nyms, synonyms, shuffled antonyms and shuffled synonyms, for each dataset we convert each pair of words $(\text{word}_1, \text{word}_2)$ into a pair of embedding vectors $(E_1, E_2)$ and then take their difference $D_{12} = E_2 - E_1$. We then project this set of difference vectors onto 2D by using the UMAP \cite{mcinnes2018umap} algorithm. There are various hyperparameters that UMAP uses, and we run variations on some of them. A critical hyperparameter for us is the metric that UMAP uses for the data to be projected: our observations only surface when we use the Euclidean metric instead of cosine similarity. We also emphasize that the UMAP algorithm is label-agnostic: we project the full dataset to 2D and color the differently labeled points after the fact.

\section{The Swirl}

\begin{figure}[htbp]
  \centering
  \includegraphics[width=\textwidth]{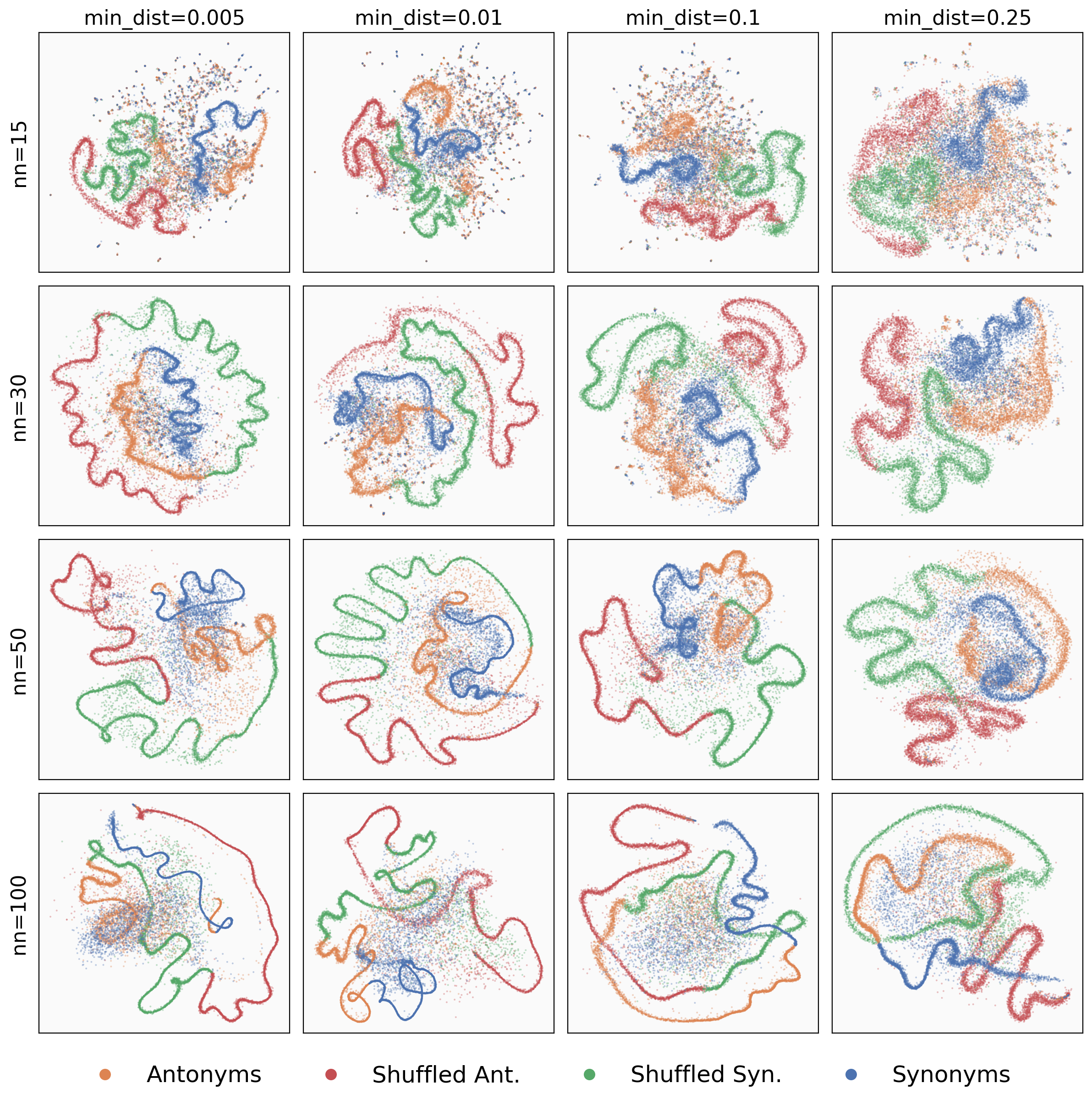}
  \caption{UMAP projections of the difference vectors from the BERT model with various hyperparameters.}
  \label{fig:umap-hp-grid-bert}
\end{figure}

Our core observation is that there seems to be a consistent swirl in the UMAP projections of the difference vectors. 
Figure \ref{fig:umap-hp-grid-bert} shows the UMAP projections of the BERT data under various UMAP hyperparameters.

There are several observations we can immediately make on Figure \ref{fig:umap-hp-grid-bert}.
\begin{enumerate}
  \item The projections have a similar `swirl' or a loop in all of the parametrizations, though there are various amounts of noise as well. To us this implies that this is not a simple projection artefact.
  \item Each of the four datasets; antonyms, synonyms, shuffled antonyms and shuffled synonyms, seems to have a strong one-dimensional structure in this projection.
  \item In the images, the blue and the green parts stay mostly separate, as do the red and the orange. This means that the antonym points are quite separated from shuffled antonyms, and synonyms from shuffled synonyms.
\end{enumerate}

\begin{figure}[htbp]
  \centering
  \includegraphics[width=\textwidth]{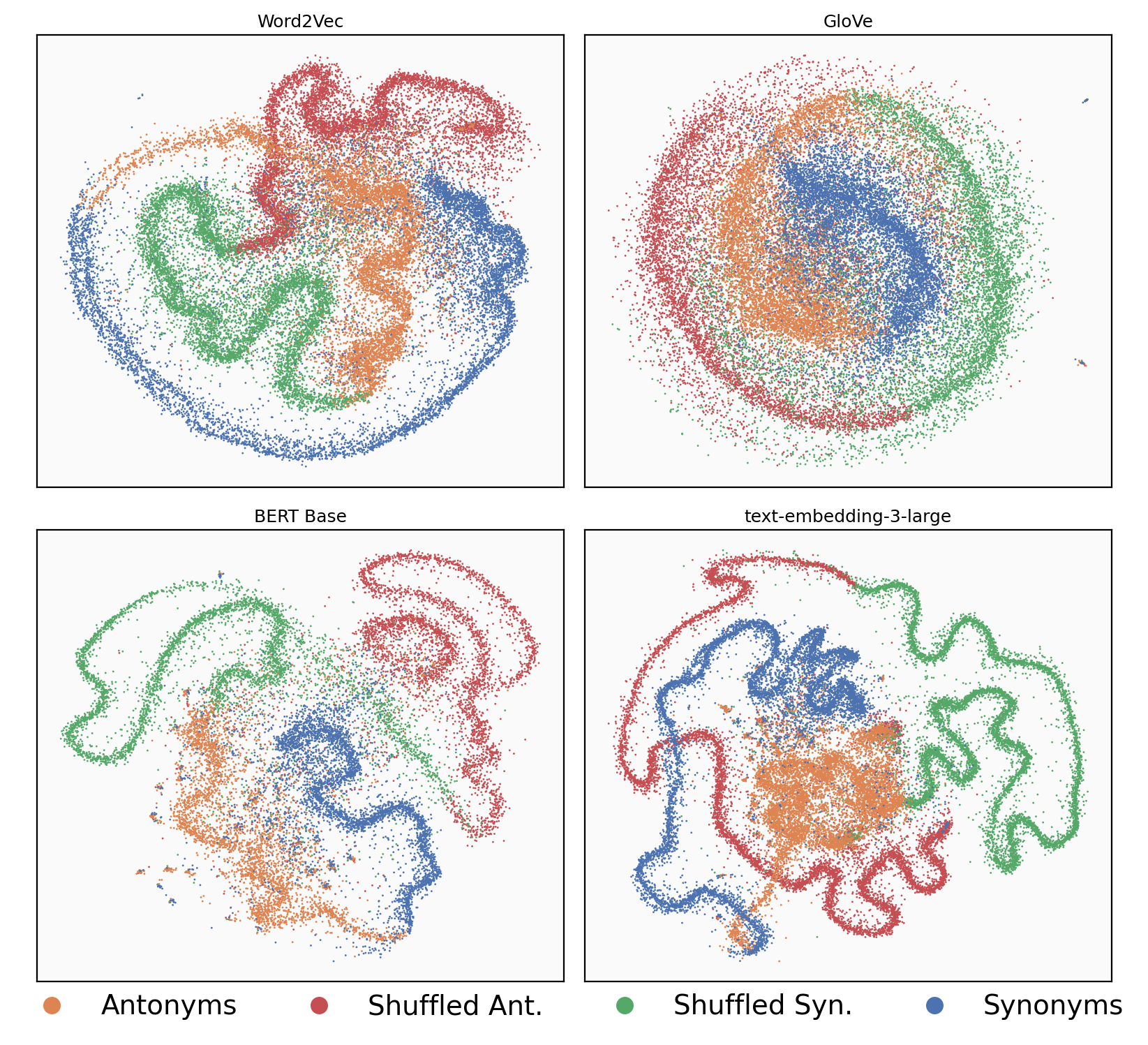}
  \caption{UMAP projections of difference vectors ($\text{\texttt{n\_neighbors}}=30$, $\text{\texttt{min\_dist}}=0.1$) for four embedding models.}
  \label{fig:umap-grid-unfiltered}
\end{figure}

This effect does not seem to be just a curiosity of the BERT context-free embeddings; Figure \ref{fig:umap-grid-unfiltered} shows the projections with the same UMAP hyperparameters of our datasets under four different models.

\subsection{An ablation of the visualization}

\begin{figure}[htbp]
  \centering
  \includegraphics[width=\textwidth]{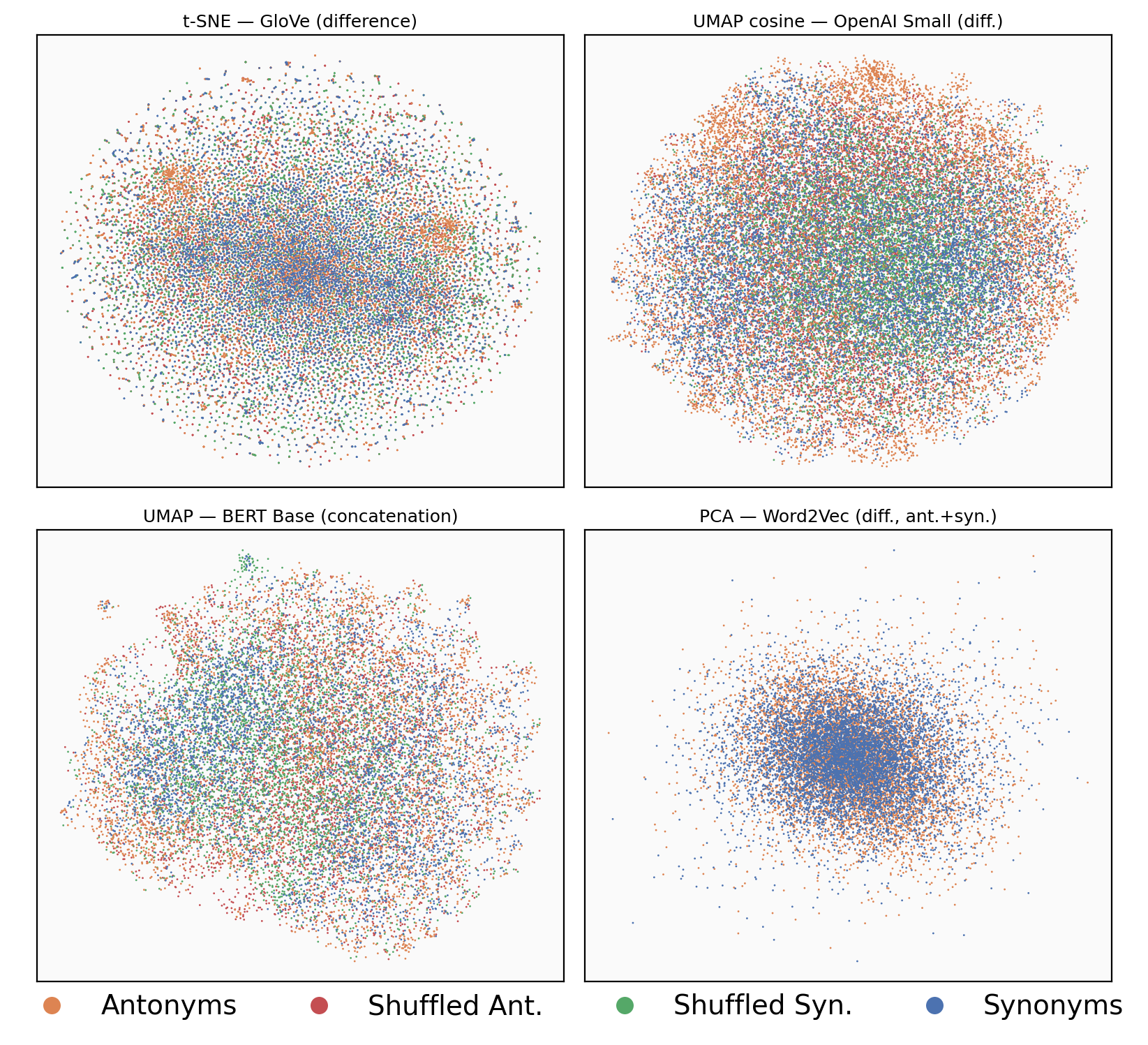}
  \caption{Projections that do \emph{not} produce the swirl pattern. Top-left: t-SNE of GloVe difference vectors. Top-right: UMAP with cosine distance (text-embedding-3-small). Bottom-left: UMAP of concatenation vectors (BERT). Bottom-right: PCA of Word2Vec difference vectors (antonyms and synonyms only).}
  \label{fig:nonresults-grid}
\end{figure}

What makes this `swirl' curious is that it appears in a somewhat stable fashion with various UMAP hyperparametrizations of difference vectors and across different models, but completely disappears if the approach is varied. In Figure \ref{fig:nonresults-grid} we show projections where we have altered some aspect of the approach; the projection method (UMAP, t-SNE or PCA), the underlying metric (euclidean or cosine in UMAP) or vector construction (difference or concatenation). This effect persisted even when varying the hyperparameters and suggests a trend.

\begin{figure}[htbp]
  \centering
  \includegraphics[width=\textwidth]{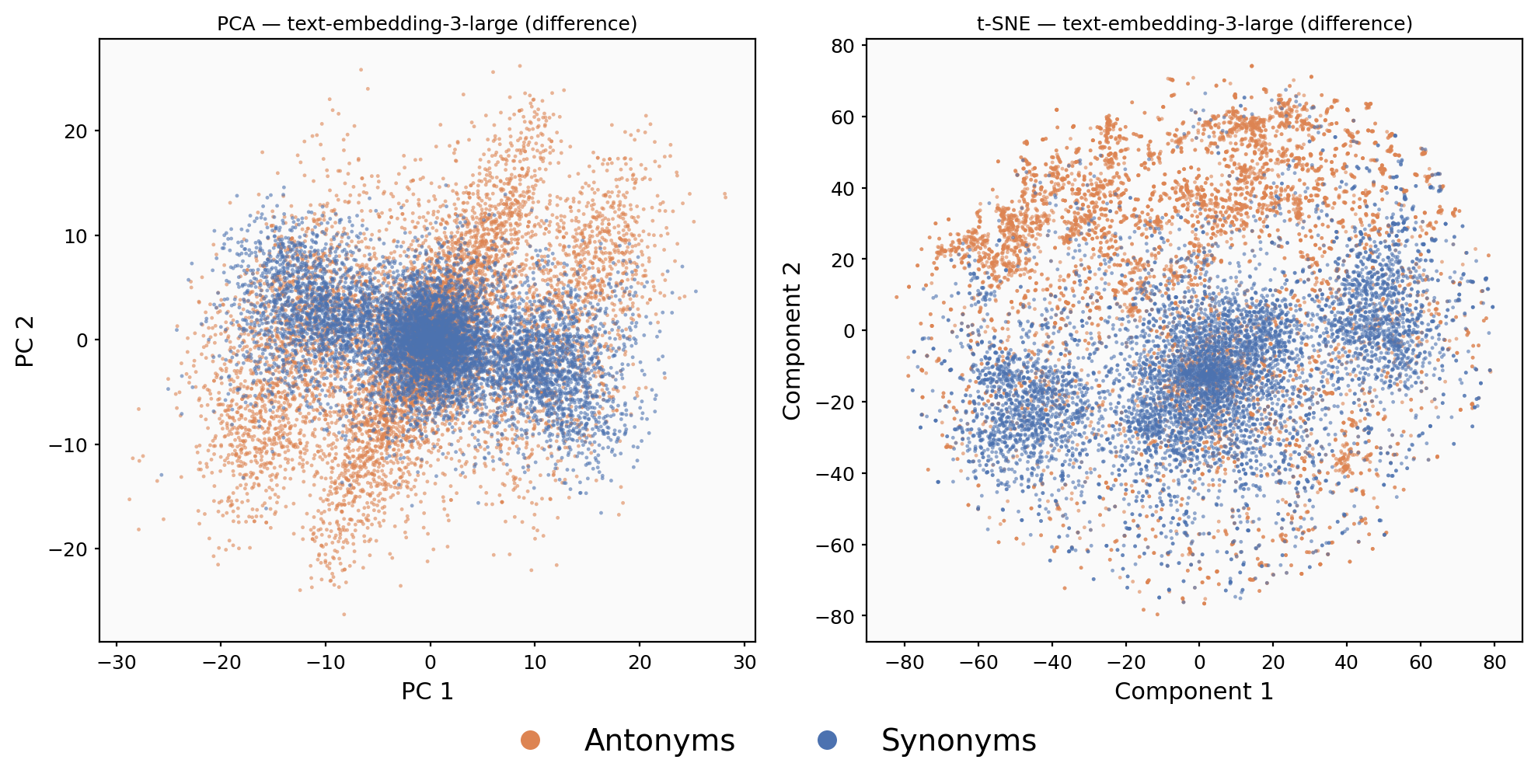}
  \caption{text-embedding-3-large with non-UMAP-difference projections, showing antonyms and synonyms. Left: PCA of difference vectors. Right: t-SNE of difference vectors.}
  \label{fig:oai-large-other-projections}
\end{figure}

The exception to this trend was text-embedding-3-large where nontrivial shapes were shown also in PCA and t-SNE; see Figure~\ref{fig:oai-large-other-projections}.

\begin{figure}[htbp]
  \centering
  \includegraphics[width=\textwidth]{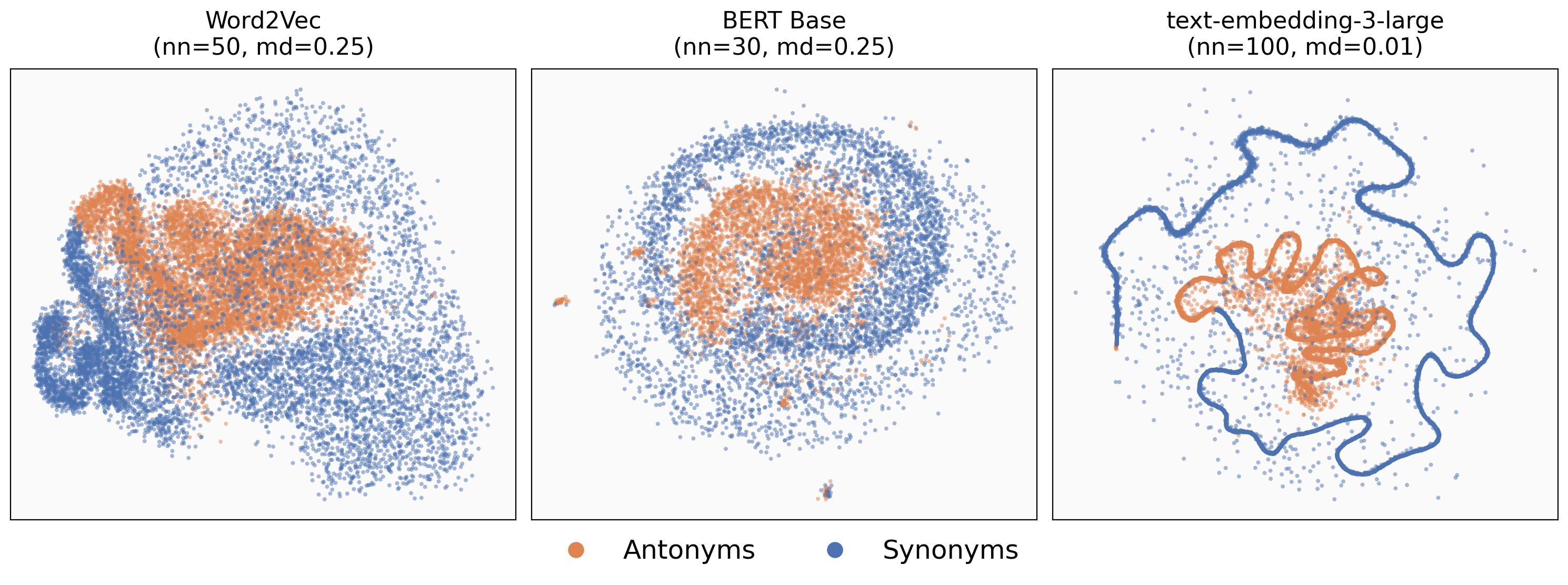}
  \caption{UMAP projections using only antonym and synonym pairs (no shuffled controls). Left: Word2Vec ($nn=50$, $md=0.25$). Center: BERT Base ($nn=30$, $md=0.25$). Right: text-embedding-3-large ($nn=100$, $md=0.01$). Without the shuffled datasets, Word2Vec and BERT show a more diffuse structure, while text-embedding-3-large retains a clear separation.}
  \label{fig:antsyn-only-selected}
\end{figure}

Also the inclusion of the shuffled datasets was somewhat crucial for the swirl to appear. Even without the shuffled datasets included the projection images tend to contain some sort of spiral-like structure, but the images tend to get much noisier and the linear components turn more diffuse. The exception here is, again, the text-embedding-3-large model, where the image stays `crisp'; see Figure \ref{fig:antsyn-only-selected} for visualizations.

\subsection{Classification}
\label{sec:classification}

Since the UMAP projections of the difference vectors have such visually striking shape, it is natural to try and classify points by the following algorithm.
\begin{itemize}
  \item Take the train and test set of word pairs. Create the shuffled word pairs.
  \item Create the difference vector point cloud for all data and UMAP project it to the plane.
  \item Remove the shuffled datasets from the image.
  \item Cluster the point cloud based on standard clustering methods.
  \item In each cluster, have the train set vote the label of that cluster.
  \item Label the test set points based on cluster inclusion.
\end{itemize}

We note that this method is not inductive but \emph{transductive}: we do not use our train data to create a model which can be applied to test points one by one. Instead we infer the labels of the test set all at once and while using the structure of the whole test set as an extra piece of information. In particular, we do not claim that the results are directly comparable to inductive methods, though we do provide SOTA inductive method results below for context.

In the ASD task literature we typically see two ways of splitting the data into train, validation and test sets. One approach is to simply use the split present in the Stuttgart dataset. However, this split does the splitting at the level of word pairs, and some words can then appear both in the train and test sets as parts of different pairs. Thus we also split the data in a ``lexical'' way where no single word was part of both a train set word pair and test set word pair. In Table \ref{table:literature_classification} we list some of the ASD task results in the past years for both types of dataset splits.

\begin{table}[ht]
  \centering
  \caption{Literature baselines by paper and year (split shown per row: Random/Stuttgart or Lexical)}
  \label{table:literature_classification}
  \resizebox{\linewidth}{!}{
  \begin{tabular}{l l c c c c c}
  \toprule
  Split setting & Paper / Method & Year & Adj F1 & Verb F1 & Noun F1 & Overall F1 \\
  \midrule
  Random / Stuttgart & AntSynNET (FastText) & 2017 & 0.773 & 0.768 & 0.817 & 0.786 \\
  Random / Stuttgart & Parasiam (FastText) & 2019 & 0.856 & 0.891 & 0.848 & 0.865 \\
  Random / Stuttgart & Distiller (GloVe) & 2019 & 0.884 & 0.891 & 0.844 & 0.873 \\
  Random / Stuttgart & MoE-ASD (FastText) & 2021 & 0.892 & 0.908 & 0.869 & 0.890 \\
  Random / Stuttgart & ICE-NET (FastText) & 2024 & 0.908 & 0.915 & 0.883 & 0.902 \\
  Random / Stuttgart & Distiller (dLCE) & 2019 & 0.928 & 0.921 & 0.911 & 0.920 \\
  Random / Stuttgart & MoE-ASD (dLCE) & 2021 & 0.938 & 0.929 & 0.935 & 0.934 \\
  Random / Stuttgart & \best{ICE-NET (dLCE)} & 2024 & 0.940 & 0.933 & 0.939 & \best{0.937} \\
  \midrule
  Lexical & Parasiam (FastText) & 2019 & 0.769 & 0.719 & 0.748 & 0.745 \\
  Lexical & MoE-ASD (FastText) & 2021 & 0.809 & 0.753 & 0.776 & 0.779 \\
  Lexical & ICE-NET (FastText) & 2024 & 0.815 & 0.758 & 0.793 & 0.789 \\
  Lexical & Parasiam (dLCE) & 2019 & 0.850 & 0.819 & 0.876 & 0.848 \\
  Lexical & MoE-ASD (dLCE) & 2021 & 0.892 & 0.847 & 0.890 & 0.876 \\
  Lexical & \best{ICE-NET (dLCE)} & 2024 & 0.898 & 0.859 & 0.900 & \best{0.886} \\
  \bottomrule
  \end{tabular}
  }
  \end{table}

In Tables \ref{table:classification_stuttgart} and \ref{table:classification_lexical} we list the results of the UMAP based transductive clustering system with various clustering methods. We also include the accuracies achieved if the UMAP projection part was omitted and the classification approach was applied to the raw embedding vectors.
  
  \begin{table}[ht]
  \centering
  \caption{Transductive inference results on the lexical data split.}
  \label{table:classification_lexical}
  \resizebox{\linewidth}{!}{
  \begin{tabular}{lcccccccc}
  \toprule
  Dataset & \multicolumn{4}{c}{Non-UMAP} & \multicolumn{4}{c}{UMAP} \\
  \cmidrule(lr){2-5}\cmidrule(lr){6-9}
   & LR & ShallowNN & KMeans & Spectral & LR & ShallowNN & KMeans & Spectral \\
  \midrule
  word2vec & 0.4936 & 0.6474 & 0.6047 & 0.4708 & 0.7619 & \best{0.9095} & \best{0.8904} & \best{0.8998} \\
  glove & 0.4993 & 0.6350 & 0.5753 & 0.4768 & 0.7183 & 0.8477 & 0.8284 & 0.8269 \\
  BERT (bert-base-cased) & 0.4653 & 0.6975 & 0.5248 & 0.3477 & 0.8562 & \best{0.9281} & \best{0.9113} & \best{0.8950} \\
  text-embedding-3-small & 0.5174 & 0.7961 & 0.7017 & 0.6537 & \best{0.9017} & \best{0.9302} & \best{0.9249} & \best{0.9249} \\
  text-embedding-3-large & 0.5232 & 0.8464 & 0.6580 & 0.5738 & \best{0.9069} & \best{0.9660} & \best{0.9643} & \best{0.9643} \\
  \bottomrule
  \end{tabular}
  }
  \end{table}
  
  \begin{table}[ht]
  \centering
  \caption{Transductive inference results on the Stuttgart data split.}
  \label{table:classification_stuttgart}
  \resizebox{\linewidth}{!}{
  \begin{tabular}{lcccccccc}
  \toprule
  Dataset & \multicolumn{4}{c}{Non-UMAP} & \multicolumn{4}{c}{UMAP} \\
  \cmidrule(lr){2-5}\cmidrule(lr){6-9}
   & LR & ShallowNN & KMeans & Spectral & LR & ShallowNN & KMeans & Spectral \\
  \midrule
  word2vec & 0.4992 & 0.8121 & 0.6064 & 0.5743 & 0.8247 & 0.9129 & 0.9122 & 0.9076 \\
  glove & 0.5056 & 0.8153 & 0.5517 & 0.5869 & 0.7453 & 0.8515 & 0.8568 & 0.8587 \\
  BERT (bert-base-cased) & 0.5025 & 0.8374 & 0.5799 & 0.3762 & 0.8414 & 0.9178 & 0.9167 & 0.8998 \\
  text-embedding-3-small & 0.4824 & 0.8845 & 0.6995 & 0.7118 & 0.9333 & 0.9209 & 0.9333 & 0.9346 \\
  text-embedding-3-large & 0.4660 & 0.9088 & 0.6863 & 0.6074 & \best{0.9432} & \best{0.9721} & 0.9358 & \best{0.9384} \\
  \bottomrule
  \end{tabular}
  }
  \end{table}

Immediately visible in Tables \ref{table:classification_stuttgart} and \ref{table:classification_lexical} is that the UMAP projection improves accuracies across the board. We also note that the text-embedding-3-large dominates in the results, even providing good classification scores without the UMAP projection.

\section{Final thoughts}

Is this ``swirl'' just a projection artefact or a result from (unintentional) $p$-hacking? We argue not. The fact that the effect is stable across three very different types of embeddings; word2vec/GloVe, BERT context-free embeddings, text-embedding-3-small/large suggests to us that there is indeed some structure at play. We do not have an active hypothesis on why exactly we see this kind of a structure in the data, but feel that it should lead to some insights about the geometry of the embeddings of antonyms, and perhaps function as an interesting example of the differences of UMAP versus other projection methods.
Finally, we note that there might be a connection to the tentatively proposed universal structures that might emerge in (L)LMs; \cite{huh2024platonic,jha2025harnessing}.

\bigskip

\textbf{Acknowledgements:} I thank my research group(s) for commentary and feedback on ``Rami's colourful swirls''. I especially appreciate the many comments that many of the swirls deserve a gallery exhibition.

\bigskip

\textbf{Statement on the usage of AI:} Coding assistants have been used extensively in this analysis. The text of this document is completely human-written, though we've at times asked for feedback from various LLMs. The scientific merits here, if any, rest on the author and I take full responsibility on the content of this work.

\bibliographystyle{alpha}
\bibliography{antonyms_bib}
\label{lastpage}
\end{document}